\documentclass[sigconf]{aamas}

\usepackage{balance} 

\setcopyright{ifaamas}
\acmConference[AAMAS '26]{Proc.\@ of the 25th International Conference
on Autonomous Agents and Multiagent Systems (AAMAS 2026)}{May 25 -- 29, 2026}
{Paphos, Cyprus}{C.~Amato, L.~Dennis, V.~Mascardi, J.~Thangarajah (eds.)}
\copyrightyear{2026}
\acmYear{2026}
\acmDOI{}
\acmPrice{}
\acmISBN{}

\usepackage{multirow} 
\usepackage{mathtools}
\usepackage{graphicx}
\usepackage{float}
\usepackage{wrapfig} 
\usepackage{algorithm} 
\usepackage{algpseudocode} 
\usepackage{booktabs} 
\usepackage{xcolor} 
\definecolor{color_links}{RGB}{0,102,204}

\usepackage{adjustbox}
\usepackage{ifthen}

\newboolean{showcomments}
\setboolean{showcomments}{true}

\ifthenelse{\boolean{showcomments}}
  {\newcommand{\nb}[3]{
  {\color{#2}\small\fbox{\bfseries\sffamily\scriptsize#1}}
  {\color{#2}\sffamily\small$\triangleright~$\textit{\small #3}$~\triangleleft$}
  }
  }
  {\newcommand{\nb}[3]{}
  }

\usepackage{colortbl,xcolor}
\usepackage{xparse}   
\definecolor{green_for_words}{RGB}{70,200,84}   
\definecolor{topone}{RGB}{169,255,119}   
\definecolor{toptwo}{RGB}{118,154,255}   
\definecolor{topthree}{RGB}{220,50,32} 
\newcommand{\colresult}[3]{%
  \begingroup
  \ifnum#3=1
    \cellcolor{topone!40}%
  \else\ifnum#3=2
    \cellcolor{toptwo!40}%
  \else\ifnum#3=3
    \cellcolor{topthree!40}%
  \fi\fi\fi
  #1$\pm$#2%
  \endgroup
}

\acmSubmissionID{792}
\title[AAMAS-2026 Formatting Instructions]{Memory Retention Is Not Enough to Master Memory Tasks in Reinforcement Learning}

\author{Oleg Shchendrigin}
\affiliation{%
  \institution{MIRIAI}
  \city{Moscow}
  \country{Russia}
}
\affiliation{%
  \institution{Innopolis University}
  \city{Innopolis}
  \country{Russia}
}
\email{shchendrigin.o@miriai.org}

\author{Egor Cherepanov}
\affiliation{%
  \institution{MIRIAI}
  \city{Moscow}
  \country{Russia}
}
\affiliation{%
  \institution{Cognitive AI Systems Lab}
  \city{Moscow}
  \country{Russia}
}
\email{cherepanov.e@miriai.org}

\author{Alexey K.\ Kovalev}
\affiliation{%
  \institution{MIRIAI}
  \city{Moscow}
  \country{Russia}
}
\affiliation{%
  \institution{Cognitive AI Systems Lab}
  \city{Moscow}
  \country{Russia}
}

\author{Aleksandr I.\ Panov}
\affiliation{%
  \institution{MIRIAI}
  \city{Moscow}
  \country{Russia}
}
\affiliation{%
  \institution{Cognitive AI Systems Lab}
  \city{Moscow}
  \country{Russia}
}

\makeatletter
\let\AAMAS@orig@mkauthors\@mkauthors
\newcommand{\CustomAuthorBlock}{%
  {\large\bfseries
    Oleg Shchendrigin\textsuperscript{1,3}\hspace{1.2em}%
    Egor Cherepanov\textsuperscript{2,3}\hspace{1.2em}%
    Alexey K.\ Kovalev\textsuperscript{2,3}\hspace{1.2em}%
    Aleksandr I.\ Panov\textsuperscript{2,3}\par
  }%
  \vspace{0.25em}%
  {\normalsize
    \textsuperscript{1} Innopolis University, Innopolis, Russia\hspace{1.5em}%
    \textsuperscript{2} Cognitive AI Systems Lab, Moscow, Russia\par
    \textsuperscript{3} MIRIAI, Moscow, Russia\par
  }%
  \vspace{0.1em}%
  {\normalsize\ttfamily
    \{shchendrigin.o, cherepanov.e\}@miriai.org\par
  }%
  \vspace{0.1em}%
  \centerline{\textbf{Project Page:} \href{https://quartz-admirer.github.io/Memory-Rewriting/}{\textcolor{color_links}{\textbf{quartz-admirer.github.io/Memory-Rewriting/}}}}%
}
\def\@mkauthors{%
  \if@ACM@anonymous
    \AAMAS@orig@mkauthors
  \else
    \begingroup
      \hsize=\textwidth
      \global\setbox\mktitle@bx=\vbox{\noindent
        \unvbox\mktitle@bx\par\medskip
        \centering
        \CustomAuthorBlock
        \par\bigskip
      }%
    \endgroup
  \fi
}

\makeatother

\begin{abstract}
Effective decision-making in the real world depends on memory that is both stable and adaptive: environments change over time, and agents must retain relevant information over long horizons while also updating or overwriting outdated content when circumstances shift. 
Existing Reinforcement Learning (RL) benchmarks and memory-augmented agents focus primarily on retention, leaving the equally critical ability of memory rewriting largely unexplored. 
To address this gap, we introduce a benchmark that explicitly tests continual memory updating under partial observability, i.e. the natural setting where an agent must rely on memory rather than current observations, and use it to compare recurrent, transformer-based, and structured memory architectures. 
Our experiments reveal that classic recurrent models, despite their simplicity, demonstrate greater flexibility and robustness in memory rewriting tasks than modern structured memories, which succeed only under narrow conditions, and transformer-based agents, which often fail beyond trivial retention cases. 
These findings expose a fundamental limitation of current approaches and emphasize the necessity of memory mechanisms that balance stable retention with adaptive updating. 
Our work highlights this overlooked challenge, introduces benchmarks to evaluate it, and offers insights for designing future RL agents with explicit and trainable forgetting mechanisms. Code: \url{https://quartz-admirer.github.io/Memory-Rewriting/}
\end{abstract}



\newcommand{\BibTeX}{\rm B\kern-.05em{\sc i\kern-.025em b}\kern-.08em\TeX}

\begin{document}


\pagestyle{fancy}
\fancyhead{}


\maketitle 


\section{Introduction}
\begin{figure}[t]
    \centering
    \includegraphics[width=1\linewidth]{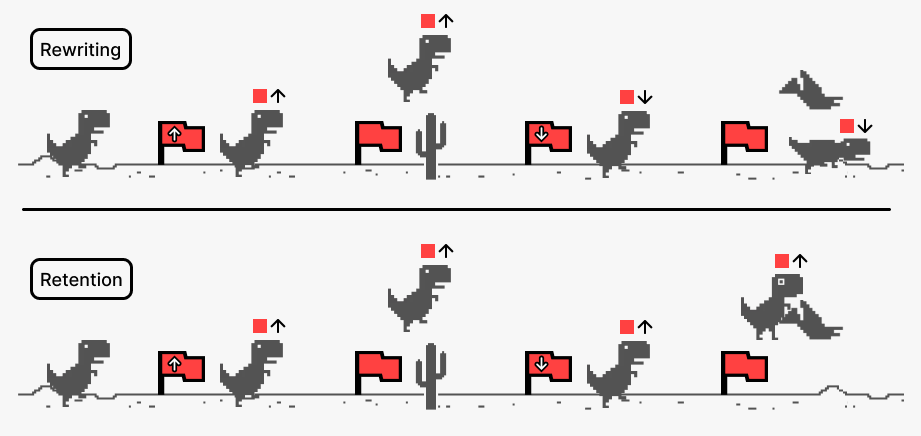}
    \vspace{-2em}
    \caption{
    \textbf{Illustration of memory rewriting vs.\ retention.} 
    The agent first observes a cue, stores it in memory, and acts according to that cue (jump for $\uparrow$). 
    Later, a new cue appears indicating a change in required behavior ($\downarrow$).  
    In the rewriting case (top), the agent overwrites its previous memory with the new cue and successfully adapts its action.  
	    In the retention case (bottom), the agent fails to update its memory, continues to act based on the old cue, and consequently performs the wrong action, leading to failure.
	    }
	    \Description{A schematic comparing two cases: memory rewriting (top) where an agent updates its stored cue after a new cue appears and succeeds, versus retention-only (bottom) where the agent keeps the old cue and fails.}
	\label{fig:visual-abstract}
	\vspace{-1em}
\end{figure}

Many everyday tasks demand memory that can both preserve and revise information~\citep{koslov2019cognitive,sakon2022differences,bretton2024suppressing}. 
A pedestrian following a sequence of directional signs must replace an earlier instruction (\emph{``turn left at the park''}) when a new sign ahead indicates a detour, 
while a warehouse robot tracking object locations must update its internal map each time an item is moved. 
In both cases, previously stored information becomes misleading unless the agent rewrites its memory with current observations 
(see Figure~\ref{fig:visual-abstract} for a representative illustration).

Reinforcement Learning (RL) agents must exhibit the same capacity to revise stored information in order to act reliably in dynamic and partially observable environments. 
In realistic settings, observations are often incomplete or ambiguous: sensors are noisy, observations are occluded (limited field of view, temporary blockages), and the latent state cannot be inferred from a single frame~\citep{lauri2022partially,meng2021memory,Kurniawati2022PartiallyOM,Zhang2023ARO}. 
Under partial observability, effective decision-making requires agents to construct and maintain a memory that summarizes the relevant aspects of past experience~\citep{parisotto2020stabilizing,cherepanov2023recurrent,le2024stablehadamardmemoryrevitalizing,Tao2025BenchmarkingPO,koepernik2025general,cherepanov2025elmur,zelezetsky2025re}.

Memory in RL is not a single ability but a collection of interdependent processes: what information to encode, how long to retain it, and how to update or discard it~\citep{ye2020retrieval,mujawar2021memory,aljundi2018memory}. 
Most existing work focuses on \emph{retention} -- storing a cue and preserving it across time~\citep{esslinger2022deep,ni2023transformers}. 
However, many decision-making problems depend equally on \emph{rewriting}: selectively discarding outdated content and integrating new evidence while avoiding interference from previously stored but now-irrelevant representations. 
Agents that only retain information risk acting on obsolete internal states. 
Despite its centrality in natural cognition, this ability to rewrite memory remains underexplored in artificial agents. 
Current benchmarks typically reward remembering a single cue until the episode ends, offering little incentive to assess continual, selective updating~\citep{lampinen2021towards,cherepanov2025memory}.

In this work, we isolate and systematically investigate \emph{memory rewriting} in RL through four diagnostic tasks that enforce continual memory updates under partial observability.  
\texttt{\textbf{Endless T-Maze}} consists of sequential corridors where each new cue invalidates the previous one, requiring the agent to actively overwrite outdated instructions.  
\texttt{\textbf{Color-Cubes}} (in \texttt{Trivial}, \texttt{Medium}, and \texttt{Extreme} variants) is a grid-world task where colored cubes stochastically teleport, forcing agents to continually update their internal map and avoid acting on stale information.

We evaluate three representative families of memory-augmented RL agents -- recurrent policies, transformer-based architectures, and structured external memories -- providing a detailed characterization of their respective strengths, limitations, and generalization patterns. 
We focus on how each architecture adapts when memory rewriting becomes essential for success, exposing distinct strategies and failure modes across these paradigms.

\textbf{Our contributions are threefold:}
\begin{enumerate}
    \item \textbf{Benchmarks for rewriting.} We introduce \texttt{Endless T-Maze} and \texttt{Color-Cubes}, two families of environments, which together provide four different options for isolate the ability to perform continual, selective memory updates beyond simple cue retention.
    \item \textbf{Systematic evaluation.} We compare recurrent, transformer, and structured-memory agents, identifying when rewriting mechanisms succeed or fail and how performance degrades with increasing memory update frequency.
    \item \textbf{Design principles.} We analyze architectural factors linked to rewriting competence, highlighting the effectiveness of \emph{explicit, adaptive forgetting} (e.g., learnable forget gates) over cached-state or rigidly structured memories, and outline practical guidelines for evaluating memory mechanisms in partially observable tasks.
\end{enumerate}

\section{Background}
\paragraph{\textbf{Markov Decision Processes.}}
A fully observable decision-making problem can be formalized as a \emph{Markov Decision Process} (MDP), 
$\mathcal{M} = (\mathcal{S}, \mathcal{A}, P, r, \mu_0, \gamma)$, 
where $\mathcal{S}$ and $\mathcal{A}$ are the state and action spaces, 
$P(\cdot\!\mid\!s,a)$ is the transition kernel, 
$r(s,a)$ the reward, $\mu_0$ the initial-state distribution, and $\gamma$ the discount factor~\citep{sutton1998reinforcement}. 
The Markov property implies that the optimal policy depends only on the current state, $\pi^\star(a_t \mid s_t)$, 
with the objective of maximizing the expected discounted return 
$\mathbb{E}\!\left[\sum_t \gamma^t r(s_t, a_t)\right]$.

\paragraph{\textbf{Partially Observable Markov Decision Processes.}}
In many real-world settings, the state $x_t$ is not directly observable. 
A \emph{Partially Observable MDP} (POMDP) extends the MDP by introducing an observation space $\mathcal{O}$ and observation kernel 
$O(o \mid s', a)$, defining the tuple 
$\mathcal{P} = (\mathcal{S}, \mathcal{A}, \mathcal{O}, P, O, r, \mu_0, \gamma)$~\citep{kaelbling1998pomdp}. 
Here, policies condition on the full interaction history 
$h_t = (o_0, a_0, \dots, o_t)$, resulting in $\pi(a_t \mid h_t)$. 
Because exact Bayesian filtering~\citep{ahmadi2020control} is typically intractable, agents approximate it using a learned memory state 
$m_t = f_\phi(h_t)$, which summarizes past information for decision-making via $\pi_\theta(a_t \mid m_t)$. 
In the fully observable limit, $m_t$ coincides with the true state, recovering the standard MDP.

\paragraph{\textbf{Dynamic Memory and Rewriting.}}
Under partial observability, agents must not only retain but also \emph{rewrite} memory as task-relevant information changes. 
Let $m_t \in \mathcal{M}$ denote the memory state and $\eta_t = (a_t, o_{t+1})$ the new input. 
A general update rule is 
$m_{t+1} = U_\phi(m_t, \eta_t)$, 
which can be decomposed as
\begin{equation}
m_{t+1} = W_\phi\!\big(F_\phi(m_t),\, E_\phi(\eta_t)\big),
\end{equation}
where $F_\phi$ selects or forgets parts of the old memory, 
$E_\phi$ encodes new input, and $W_\phi$ integrates both. 
Rewriting thus represents a selective update process -- information is preserved, attenuated, or replaced depending on current relevance -- ensuring that the memory state evolves to maintain only the most decision-relevant features of experience.

\paragraph{\textbf{Memory-intensive environments and dynamic correlations.}}
A partially observable environment can be described by a set of \emph{correlation horizons} 
$\Xi = \{\xi_n\}$, where each $\xi = t_r - t_e - \Delta t + 1$ measures the temporal gap between an informative event $\alpha_{t_e}^{\Delta t}$, which begins at time $t_e$ and lasts for $\Delta t$ steps, and the later decision $\beta_{t_r}$ made at time $t_r$ that depends on it~\citep{cherepanov2024unraveling}. 
An environment is \emph{memory-intensive} when $\min \Xi > 1$, requiring the agent to recall information across multiple timesteps. 
In our tasks, the events $\alpha_{t_e}^{\Delta t}$ are \emph{dynamic} and evolve over time ($\alpha_{t_e}^{\Delta t} = \alpha_{t_e}^{\Delta t}(t)$), producing shifting and overlapping correlation horizons. 
This dynamic structure forces the agent to not only retain long-range dependencies but also to rewrite outdated information as conditions change—an ability explicitly probed by our \texttt{Endless T-Maze} and \texttt{Color-Cubes} environments.

\section{Related Work}
\subsection{Memory-augmented RL agents.}
Partial observability is a defining property of many real-world decision-making problems, where agents must infer hidden state information from incomplete and noisy observations. 
To act effectively, such agents require mechanisms that can retain, integrate, and adapt information over time. 
This need has motivated a broad line of research on \emph{memory-augmented RL} architectures, which differ in how they represent, store, and update temporal dependencies.

\textbf{Recurrent architectures.} 
A common approach introduces recurrent neural networks into policy and value functions. 
The \textbf{PPO-LSTM}~\citep{schulman2017proximalpolicyoptimizationalgorithms} extends Proximal Policy Optimization (PPO,~\citep{schulman2017proximalpolicyoptimizationalgorithms}) with Long Short-Term Memory (LSTM,~\citep{Hochreiter1997LongSM}) units, 
allowing agents to maintain internal states that summarize recent experience. 
The gating mechanisms of the LSTM regulate what information is preserved or forgotten at each step, 
providing a learnable and adaptive form of memory that serves as a strong baseline for studying sequential decision-making in RL.

\textbf{Transformer-based architectures.} 
Transformers have recently been adapted to RL to leverage their ability to model long-range dependencies. 
The \textbf{Gated Transformer-XL} (\textbf{GTrXL})~\citep{parisotto2020stabilizing} extends the Transformer-XL architecture (TrXL)~\citep{dai2019transformerxlattentivelanguagemodels} for RL by introducing identity-map reordering and gating mechanisms that improve training stability. 
It preserves long-range temporal context through cached hidden states, enabling policies to leverage information beyond the immediate observation window.
However, transformers lack an explicit mechanism for selective forgetting, 
which limits their stability in long-horizon or sparse-reward environments.

\textbf{Structured memory architectures.}
To achieve more robust and interpretable forms of temporal abstraction, recent work has introduced structured external memory systems. 
The \textbf{Fast and Forgetful Memory} (\textbf{FFM},~\citep{morad2023reinforcementlearningfastforgetful}) models memory as a set of exponentially decaying traces, 
enabling gradual forgetting of outdated information inspired by computational psychology. 
The \textbf{Stable Hadamard Memory} (\textbf{SHM},~\citep{le2024stablehadamardmemoryrevitalizing}) extends this idea by learning a dynamic calibration matrix that adaptively regulates which components of memory are reinforced or suppressed, enhancing stability during online updates.

Across these architectures -- recurrent, transformer-based, and structured -- most research has focused on improving memory \emph{retention} and stability. 
In contrast, the equally important ability to \emph{rewrite} memory -- selectively discarding obsolete content and incorporating new, task-relevant information -- remains largely unexplored. 
Our work isolates this capability and provides a systematic framework for evaluating how existing memory mechanisms handle continual memory updating in partially observable settings.

\subsection{Existing benchmarks for memory testing}
A wide range of benchmarks has been developed to evaluate how RL agents store, retain, and use information under partial observability. These environments differ in sensory modality, task abstraction, and the specific aspects of memory they test.

\textbf{3D embodied environments.}  
First-person view benchmarks such as DeepMind Lab~\citep{beattie2016deepmindlab} and MiniWorld~\citep{chevalierboisvert2023minigridminiworldmodular} test navigation and spatial reasoning under occlusion, 
though agents often rely on perceptual shortcuts instead of true long-term memory. 
Memory Maze~\citep{pasukonis2022evaluatinglongtermmemory3d} enforces integration of temporally separated cues for localization, 
while ViZDoom-Two-Colors~\citep{memup} presents 3D navigation tasks where agents must remember previously observed color cues to make correct decisions later in the episode.

\textbf{2D grid-world and diagnostic tasks.}  
Grid- and vector-based environments provide controlled settings for studying temporal dependencies under partial observability.  
MiniGrid~\citep{chevalierboisvert2023minigridminiworldmodular} and Memory Gym~\citep{pleines2025memory} test sequence recall and delayed reasoning,  
while POPGym~\citep{morad2023popgymbenchmarkingpartiallyobservable} and POPGym Arcade~\citep{wang2025popgymarcadeparallelpixelated} cover both vector and visual modalities, emphasizing puzzle-like memory and POMDP control tasks.  
Synthetic POMDPs~\citep{wang2025synthetic} extend this idea by procedurally generating tasks with tunable observability and horizon length.  
The T-Maze~\citep{ni2023transformers} provides a minimal diagnostic setup where the agent must recall an early cue to act correctly later,  
and Endless Memory Gym~\citep{pleines2025memory} introduces adaptive, unbounded variants that evaluate continual memory updating over extended horizons.

\textbf{Cognitive and multi-modal benchmarks.}  
Beyond navigation and sequence recall, several benchmarks target higher-level cognitive abilities. 
DeepMind's Memory Tasks Suite~\citep{lampinen2021towards}, Animal-AI~\citep{voudouris2025animalaienvironmentvirtuallaboratory}, and BabyAI~\citep{chevalier2018babyai} test reasoning, instruction following, and object permanence.
The POBAX benchmark~\citep{tao2025benchmarking} evaluates memory in partially observable settings through \emph{memory improvability} -- the benefit of adding memory to agents -- across diverse tasks that require recalling information over time.

\textbf{Memory benchmarks in robotics.}  
Memory has also emerged as a key challenge in robotic control, where agents must reason over continuous, partially observable dynamics. 
Benchmarks such as MIKASA-Robo~\citep{cherepanov2025memory} and MemoryBench~\citep{sam2act} introduce visuomotor tasks that require recalling and updating latent scene properties across multiple interactions, emphasizing the role of memory in embodied decision-making.

Existing benchmarks mainly test memory \emph{retention} -- preserving information over time -- but rarely the ability to \emph{rewrite} it when cues become outdated.  
Our Endless T-Maze and Color-Cubes benchmarks fill this gap by requiring agents to continually discard obsolete cues and integrate new evidence under partial observability.  
Unlike Endless Memory Gym~\citep{pleines2025memory}, which evaluates how long agents can preserve growing sequences of information in unbounded tasks, our environments focus on the opposite challenge -- when and how agents should overwrite prior beliefs as task conditions change.  
This shift from testing memory \emph{capacity} to testing memory \emph{adaptivity} provides a complementary perspective on continual learning and exposes limitations in current architectures' ability to update internal representations.

\section{Benchmarking Memory Rewrite}
Evaluating memory in RL has traditionally focused on how well agents retain past information.  
However, many real-world problems demand not just stable retention but also the ability to update internal representations when previously relevant cues become obsolete.  
To isolate and measure this capability, we introduce a set of controlled diagnostic environments specifically designed to stress continual memory rewriting under partial observability.

Our framework comprises two families of tasks, \textbf{\texttt{Endless T-Maze}} and \textbf{\texttt{Color-Cubes}}, instantiated at three difficulty levels \texttt{Trivial}, \texttt{Medium}, and \texttt{Extreme}. All environments require agents to detect when internal information have become outdated and must be replaced, shifting the evaluation focus from long-term memory retention to adaptive memory manipulation. \textbf{\texttt{Endless T-Maze}} constitutes a minimal diagnostic that isolates the ability to overwrite old information with new one. At every corridor, the agent receives a fresh left/right cue that completely overrides the previous one, and only the most recent cue is relevant for the subsequent junction. In contrast, \textbf{\texttt{Color-Cubes}} serves as a full-fledged stress test for selective and context-dependent forgetting. The agent navigates a grid world to collect cubes of a target color, while non-target cubes teleport over time. State updates are revealed only after specific events, and information that was previously irrelevant can later become useful again. As a result, the agent must determine what to discard, what to retain, and what to reinstate, thereby demanding not only memory rewriting but selective maintenance under uncertainty.

We evaluate a diverse range of memory-augmented RL architectures -- including recurrent models, transformer-based agents, and structured external memory systems -- enabling systematic comparison of the strategies by which different mechanisms support continual memory updating. Taken together, these environments provide a principled platform for investigating when and how agents rewrite memory, a capacity that is essential for adaptive behavior in non-stationary and partially observable domains.

\paragraph{\textbf{Endless T-Maze.}}
This environment extends the T-Maze~\citep{ni2023transformers} into an infinite sequence of interconnected corridors, forming a continual version of the cue-based navigation task.  
At the start of each corridor, the agent receives a binary cue indicating whether to turn left or right at the upcoming junction.  
Once a turn is made, the cue changes, invalidating all previous information -- thereby requiring continual memory overwriting rather than static retention.  

The observation space is a compact vector encoding the current cue and the agent's normalized position within the corridor, while the action space includes \texttt{MOVE\_\allowbreak FORWARD}, \texttt{TURN\_\allowbreak LEFT}, and \texttt{TURN\_\allowbreak RIGHT}.  
Rewards are assigned as $+1$ for a correct turn, $-1$ for an incorrect turn, and $-0.01$ if the agent does not move forward to the junction.  
Although the task is theoretically unbounded, in experiments we fix either the number of corridors or their maximum length to control task difficulty and ensure stable training.  
We also evaluate two sampling regimes for corridor lengths:  
(i) \textbf{Fixed}, where all corridors share the same length $l_{\max}$, and  
(ii) \textbf{Uniform}, where lengths are sampled from a uniform distribution $l \sim \mathcal{U}[1, l_{\max}]$, introducing stochastic variation in cue timing and memory horizon.  
Further implementation details are provided in Appendix~\ref{app:t-maze}.

\begin{figure}[t]
    \centering
    \includegraphics[width=1\linewidth]{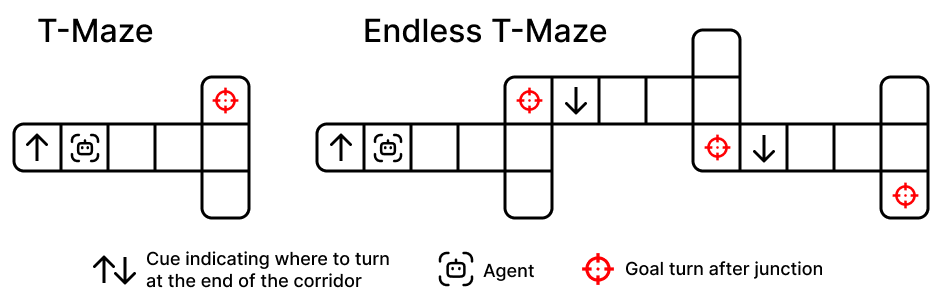}
    \vspace{-20pt}
	    \caption{\textbf{\texttt{T-Maze} vs. \texttt{Endless T-Maze}.} 
	    In the classic \texttt{T-Maze} (left), the agent receives a cue at the start and must remember it until the junction. 
	    Our proposed \texttt{Endless T-Maze} (right) extends this setup by chaining junctions: at each corridor, a fresh cue overrides all previous ones, and continual rewriting rather than simple retention is necessary to success.}
	    \Description{Two side-by-side diagrams: a classic T-Maze with one cue to remember until a junction, and an Endless T-Maze where new cues appear in each corridor and overwrite previous cues across multiple junctions.}
	    \label{fig:endless-t-maze}
\end{figure}

\begin{figure}[t]
    \centering
    \includegraphics[width=1.0\linewidth]{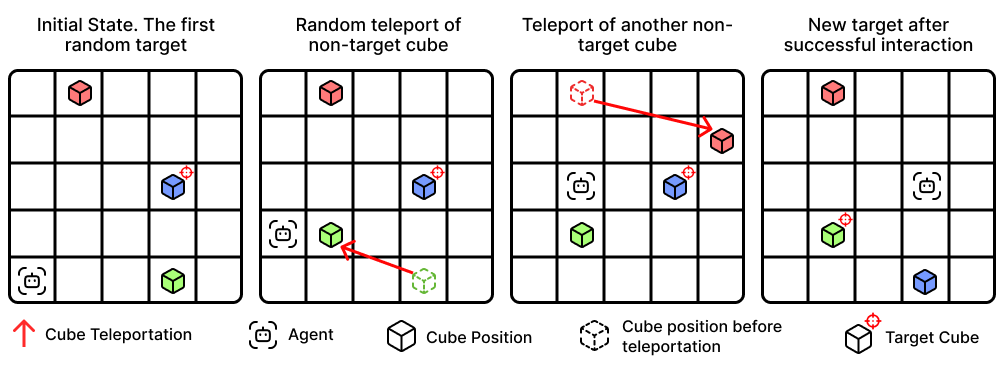}
    \vspace{-20pt}
    \caption{Visualization of key principles of the Color-Cubes environment, such as Initialization, Stochastic Teleportation, Successful Interaction.}
    \Description{A multi-panel visualization of the Color-Cubes grid world showing an initialization phase, stochastic cube teleportations during navigation, and successful interaction with a target cube.}
    \label{fig:color-cube}
\end{figure}

\paragraph{\textbf{Color-Cubes.}}
\texttt{Color-Cubes} (Figure~\ref{fig:color-cube}) is a grid-based environment that evaluates an agent's ability to maintain, detect, and update internal representations under partial observability.  
The agent acts on a $G{\times}G$ grid containing $N$ uniquely colored cubes, each assigned a color identifier $c \in \{0, 1, \dots, N{-}1\}$.  
Each episode consists of $K$ consecutive interaction phases. At the beginning of each phase, the agent is assigned a new target color $c_{\text{target}}$, indicating which cube it must locate and interact with.  
During this initialization step, the agent observes the color of every cube and the target color $c_{\text{target}}$, along with its own $(x, y)$ position and the $(x, y)$ coordinates of all cubes.  
This color information is shown only at the start of the phase or when the cube is accidentally teleported while the agent is moving toward the target, forcing the agent to rely on its memory of the target color and the initial color-position mapping to act correctly. To maintain partial visibility and not give the agent too many clues, there is a teleportation probability parameter $p_{\text{teleport}}$, which should not be set too high.

After the start of each phase, the agent must navigate to the cube with color $c_{\text{target}}$ and execute the \texttt{INTERACT} action. During the agent's movement, any cube other than the current target cube may perform a stochastic teleportation with a certain probability $p_{\text{teleport}}$, forcing the agent to detect inconsistencies between its memory and the current environment and update its internal map accordingly.
Following each successful interaction, the target color changes, and the lifted cube is teleported to a new location.
The agent receives a reward of $+1$ for a successful interaction and a small negative reward when moving away from the target cube.  
An episode terminates after successfully completing a number of phases or when the time expires.

We define three difficulty levels that progressively increase the demands on memory rewriting:

\begin{itemize}
    \item \textbf{\texttt{Trivial}} -- Single cube and single target ($N{=}K{=}1$).  
    The agent only needs to memorize the color of one cube and perform one interaction; no rewriting is required.

    \item \textbf{\texttt{Medium}} -- Multiple cubes with complete state updates (positions and colors).  
    During the agent's movement, several cubes may teleport, and the agent must update its memory to track the new configuration.

    \item \textbf{\texttt{Extreme}} -- Multiple cubes with \emph{incomplete} teleportation updates (positions only, colors hidden).  
    The agent must infer which cube moved based on prior knowledge, update its internal color–position mapping, and act accordingly, testing both rewriting and inference under uncertainty.
\end{itemize}

Further details on the environment configuration and dynamics are provided in Appendix~\ref{app:cubes}.

\section{Experiments}
We empirically evaluate how different memory architectures cope with continual memory rewriting under partial observability.
We organize our investigation around four key research questions (RQs) that collectively probe different aspects of memory rewriting:
\begin{itemize}
    \item \textbf{RQ1 (Retention).} How do different memory mechanisms behave in trivial settings where rewriting is unnecessary (e.g., $n{=}1$ \texttt{Endless T-Maze} and \texttt{Trivial} \texttt{Color-Cubes})?
    \item \textbf{RQ2 (Rewriting).} Which mechanisms can reliably rewrite (i.e., discard obsolete cues and integrate new ones) when tasks explicitly require continual updates?
    \item \textbf{RQ3 (Generalization).} How well do these mechanisms interpolate and extrapolate across horizons and rewrite frequencies in the \texttt{Endless T-Maze} environment (varying corridor lengths $l$ and counts $n$; fixed vs.\ uniform regimes)?
    \item \textbf{RQ4 (Prioritization).} When multiple updates accumulate, which mechanisms can re-rank and maintain a consistent, up-to-date memory (e.g., \texttt{Color-Cubes} \texttt{Medium}/\texttt{Extreme}) rather than merely forgetting?
\end{itemize}

\subsection{Baselines}
To identify which architectural principles support effective memory rewriting, we evaluate a broad set of agents covering both established and state-of-the-art (SOTA) memory mechanisms. 
Our study includes recurrent (\textbf{PPO-LSTM and PPO-GRU}~\citep{pleines2022generalizationmayhemslimitsrecurrent}), transformer-based (\textbf{GTrXL}~\citep{parisotto2020stabilizing}), and structured-memory architectures (\textbf{FFM}~\citep{morad2023reinforcementlearningfastforgetful}, \textbf{SHM}~\citep{le2024stablehadamardmemoryrevitalizing}).  
In addition, we use a memory-free \textbf{PPO-MLP} baseline to isolate the effect of explicit memory mechanisms under partial observability.  
Together, these agents represent the principal design paradigms for sequential decision-making in RL -- recurrent dynamics, attention-based context, and structured long-term memory -- and enable a systematic analysis of how each handles continual memory updating.

\subsection{Training and Evaluation Protocol}
We adopt a unified training and evaluation protocol to ensure comparability across agents and environments.

\paragraph{\textbf{Training configurations.}}
For \texttt{Endless T-Maze}, agents are trained across multiple combinations of corridor length $l$, number of corridors $n$, and corridor-length sampling regime (\textbf{Fixed} vs.~\textbf{Uniform}); all configurations are summarized in Table~\ref{tab:baselines}.  
For \texttt{Color-Cubes}, we train agents on the three predefined difficulty modes -- \texttt{Trivial}, \texttt{Medium}, and \texttt{Extreme} -- with detailed parameters provided in Appendix~\ref{app:cubes}.

\paragraph{\textbf{Evaluation setup.}}
For \texttt{Endless T-Maze}, each checkpoint is evaluated on the validation configurations listed in Table~\ref{tab:baselines}.  
We include both \emph{matched} settings (same $l$, $n$, and sampling regime as training) and \emph{cross-setting} evaluations to probe generalization through \emph{interpolation} (shorter corridors or fewer rewrite operations than in training) and \emph{extrapolation} (longer corridors or more rewrites).  
For \texttt{Color-Cubes}, agents trained on each difficulty mode are evaluated within that mode, allowing us to measure how well memory mechanisms handle increasing demands on continual updating and partial observability.

\begin{table}[t]
\centering
\small
\caption{
Performance of all evaluated agents on the \texttt{Endless T-Maze}.  
Each configuration is defined by the corridor length $l$ and the number of corridors $n$ under two regimes: \textbf{Fixed} (constant length) and \textbf{Uniform} (randomized length).  
\textcolor{green_for_words}{Top-1} and \textcolor{toptwo}{top-2} results are highlighted.  
Here, $n{=}1$ corresponds to the classic single-turn T-Maze (pure retention), while larger $n$ values require continual memory rewriting.
}

\label{tab:baselines}
\begin{adjustbox}{width=1\columnwidth}
\begin{tabular}{p{0.9cm}ccccccc}
\toprule
\textbf{Regime} & \textbf{$l$} & \textbf{$n$} 
 & \textbf{PPO-LSTM} 
 & \textbf{GTrXL} 
 & \textbf{SHM} 
 & \textbf{FFM} 
 & \textbf{PPO-MLP} \\
\midrule
\multirow{8}{*}{\rotatebox{90}{\textbf{Fixed}}}
 & 5  & 1  & \colresult{1.00}{0.00}{1} & 0.50$\pm$0.19 & \colresult{1.00}{0.00}{1} & \colresult{1.00}{0.00}{1} & 0.50$\pm$0.00 \\
 & 5  & 3  & \colresult{1.00}{0.00}{1} & 0.17$\pm$0.06 & \colresult{0.67}{0.24}{2} & \colresult{1.00}{0.00}{1} & 0.24$\pm$0.05 \\
 & 5  & 5  & \colresult{1.00}{0.00}{1} & 0.04$\pm$0.02 & \colresult{0.84}{0.13}{2} & \colresult{1.00}{0.00}{1} & 0.05$\pm$0.02 \\
 & 5  & 10 & \colresult{1.00}{0.00}{1} & \colresult{0.61}{0.17}{2} & 0.23$\pm$0.22 & \colresult{1.00}{0.00}{1} & 0.00$\pm$0.00 \\
 & 10 & 1  & \colresult{1.00}{0.00}{1} & 0.43$\pm$0.01 & \colresult{0.84}{0.16}{2} & \colresult{1.00}{0.00}{1} & 0.50$\pm$0.00 \\
 & 10 & 3  & \colresult{1.00}{0.00}{1} & 0.17$\pm$0.01 & \colresult{0.93}{0.07}{2} & \colresult{1.00}{0.00}{1} & 0.16$\pm$0.02 \\
 & 10 & 5  & \colresult{0.69}{0.31}{2} & 0.19$\pm$0.12 & 0.52$\pm$0.24 & \colresult{1.00}{0.00}{1} & 0.04$\pm$0.00 \\
 & 10 & 10 & \colresult{0.83}{0.17}{2} & 0.02$\pm$0.02 & 0.02$\pm$0.01 & \colresult{0.73}{0.27}{1} & 0.00$\pm$0.00 \\
\midrule
\multirow{8}{*}{\rotatebox{90}{\textbf{Uniform}}}
 & 5  & 1  & \colresult{1.00}{0.00}{1} & 0.47$\pm$0.08 & 0.55$\pm$0.23 & \colresult{0.92}{0.08}{2} & 0.26$\pm$0.02 \\
 & 5  & 3  & \colresult{1.00}{0.00}{1} & 0.15$\pm$0.09 & 0.04$\pm$0.01 & \colresult{0.43}{0.23}{2} & 0.01$\pm$0.01 \\
 & 5  & 5  & \colresult{1.00}{0.00}{1} & \colresult{0.04}{0.01}{2} & \colresult{0.04}{0.02}{2} & \colresult{0.03}{0.03}{2} & 0.00$\pm$0.00 \\
 & 5  & 10 & \colresult{1.00}{0.00}{1} & 0.00$\pm$0.00 & 0.00$\pm$0.00 & 0.00$\pm$0.00 & 0.00$\pm$0.00 \\
 & 10 & 1  & \colresult{1.00}{0.00}{1} & 0.57$\pm$0.02 & \colresult{0.91}{0.09}{2} & \colresult{1.00}{0.00}{1} & 0.24$\pm$0.12 \\
 & 10 & 3  & \colresult{0.98}{0.02}{1} & \colresult{0.16}{0.03}{2} & 0.06$\pm$0.04 & 0.00$\pm$0.00 & 0.00$\pm$0.00 \\
 & 10 & 5  & \colresult{1.00}{0.00}{1} & \colresult{0.06}{0.01}{2} & 0.04$\pm$0.01 & 0.00$\pm$0.00 & 0.00$\pm$0.00 \\
 & 10 & 10 & \colresult{1.00}{0.00}{1} & \colresult{0.01}{0.01}{2} & 0.00$\pm$0.00 & 0.00$\pm$0.00 & 0.00$\pm$0.00 \\
\bottomrule
\end{tabular}
\end{adjustbox}
\end{table}

\paragraph{\textbf{Metrics and aggregation.}}
Performance is reported as the \emph{success rate}, defined as the fraction of correctly completed trials over 100 independent evaluation episodes per checkpoint.  
For each training run, we first compute the mean success rate across these 100 episodes.  
We then report the overall mean and standard error of the mean (mean~$\pm$~SEM) across the ten independent runs.  
This aggregation captures both episode-level variability and the stability of learning across random seeds.

\paragraph{\textbf{Reproducibility and configuration control.}}
All environment parameters used for training and validation -- including $l$, $n$, and the sampling regime for \texttt{Endless T-Maze}, as well as grid size, number of cubes, and teleportation probability for \texttt{Color-Cubes} -- are detailed in Appendix~\ref{app:sec:envs}.  
Each agent was tuned individually for each environment, after which its hyperparameters were kept fixed across all training and validation configurations of that environment to ensure consistency within comparisons.  
The full list of hyperparameters and implementation details is provided in Appendix~\ref{app:training}.

\section{Results}
In order to group meaningful results across different environments and parameters, we propose to consider four research questions.

\paragraph{\textbf{RQ1: How different memory mechanisms deal with the trivial cases, where rewriting is not necessary?}}

In the simplest \texttt{Endless T-Maze} task  where $n=1$ (become classic T-Maze), which only requires an agent to write the cue once and retain it through the episode, we evaluated the baseline performance of considered models. The results are presented in Table~\ref{tab:baselines} (rows where $n=1$).

The PPO-LSTM, SHM, and FFM agents demonstrated best performance, achieving a perfect success rate ($1.00\pm0.00$) in all scenarios. In contrast, the GTrXL agent's success rates hovering around $50\%$ like MLP's, indicating it could not reliably solve the task.
In the \texttt{Trivial} \texttt{Color-Cubes} case represented in the first row in Table \ref{fig:color-cube}, the situation is slightly different. PPO-LSTM showed a result of $0.52\pm0.10$, while FFM, GTrXL, and SHM showed perfect success. 

Based on the results, we can assume that the memory mechanisms of the PPO-LSTM, SHM, and FFM architectures are capable of retaining the necessary information within the scope of our experimental protocols. A possible reason for the low success rate of GTrXL in conditions where the context length exceeds the episode length on \texttt{Endless T-Maze} is its instability in sparse reward conditions, which is confirmed by its high success rate on \texttt{Color-Cubes}, where the agent regularly receives penalties or rewards. The stability challenges of Transformer-based RL agents under sparse rewards also have been noted in prior work~\citep{chen2024transdreamerreinforcementlearningtransformer}.

\begin{figure}
    \centering
    \includegraphics[width=1.05\linewidth]{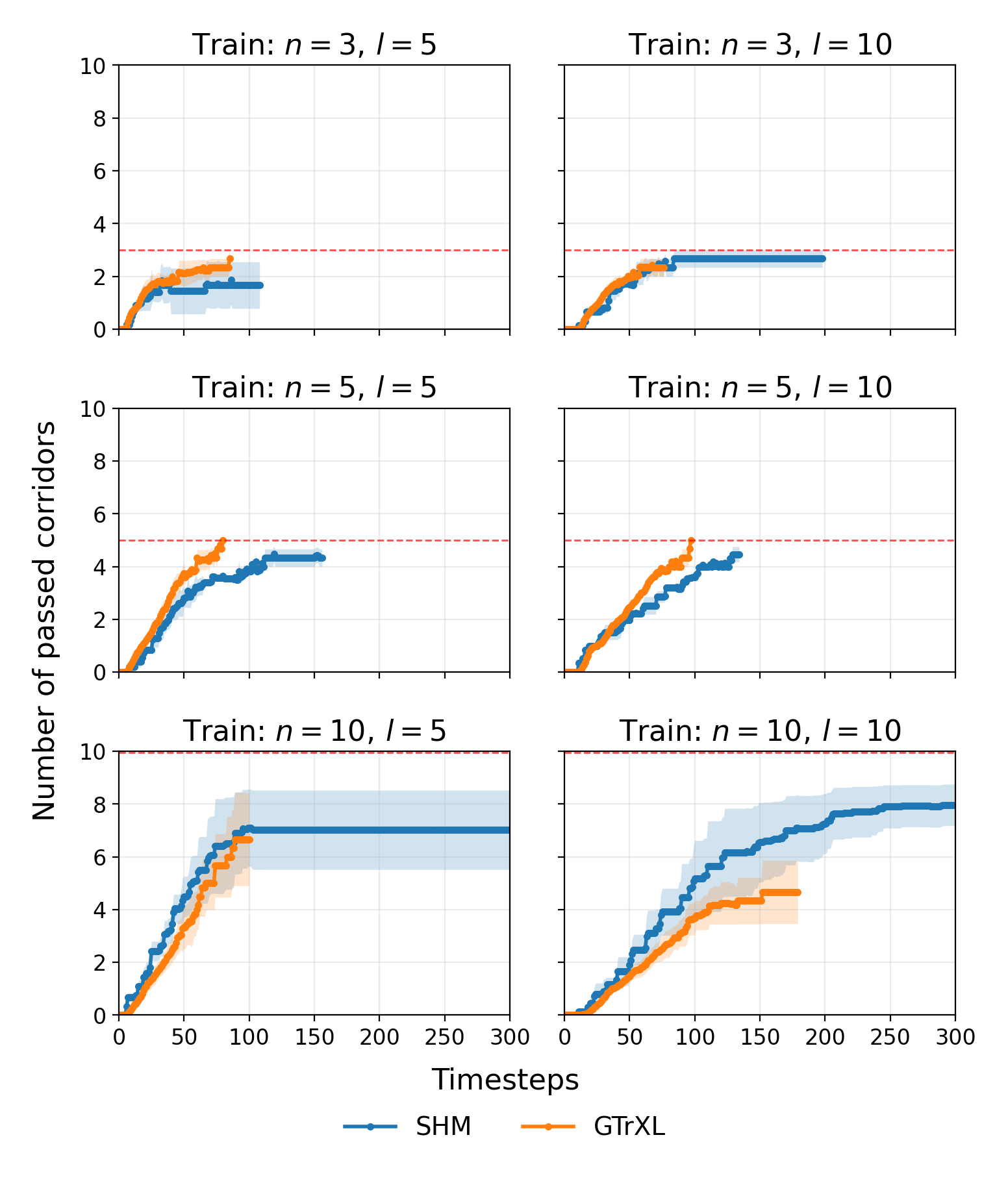}
    \vspace{-25pt}    
	    \caption{
	    \textbf{Intermediate progress of SHM and GTrXL agents in the \texttt{Endless T-Maze}.} 
	    The figure show the number of successfully passed corridors over time for different training configurations of corridors number $n$ and length $l$. Despite low overall success rates, both agents can partially  progress through several corridors before failing, indicating limited but nonzero capacity for short-term memory rewriting. Dashed red lines denote the target number of corridors for each validation configuration to reach success. 
	    }
	    \Description{A plot of partial corridor progress over time for SHM and GTrXL across multiple training configurations (varying corridor count and length), with dashed red lines marking the target corridor count for success.}
	    \label{fig:cumulative-coridors}
\end{figure}

\paragraph{\textbf{RQ2: What memory mechanisms cope with rewriting?}}
To try to identify key features of memory mechanisms that aid in rewriting we consider \texttt{Endless T-Maze} validation tasks in which the length of the corridor, the number of corridors, and the sampling were identical to the training configuration, and the length of the corridor was greater than 1 for \texttt{Endless T-Maze} environment in Table~\ref{tab:baselines}.

The PPO-LSTM agent demonstrated complete success in most \texttt{Endless T-maze} tasks considered. Meanwhile, FFM was only able to successfully cope with tasks in fixed sampling mode with $1.00\pm0.00$ result. 
SHM also demonstrated success in fixed tasks, but within 5 corridors ($n=5$) and less than FFM in configurations with $n=5$ ($0.84\pm0.13$ in $l=5$ and $0.52\pm0.24$ in $l=10$). 
GTrXL and MLP were unable to produce meaningful results in tasks that clearly required memory rewriting.

To better understand how GTrXL and SHM behave in the \texttt{Endless T-Maze}, we examine how many corridors each agent can successfully traverse within an episode despite their overall low success rates (Figure~\ref{fig:cumulative-coridors}).  
This analysis reveals how long these architectures can maintain accurate cue following before their memory representations deteriorate.  
In tasks with $n{=}3$ and $n{=}5$, both agents typically manage to navigate one corridor fewer than required, as indicated by the interrupted curves showing partial completion of the maze.

These results reveal that the baselines ability to handle memory rewriting is highly dependent on environmental predictability. The mechanism in PPO-LSTM is robust enough to succeed in both predictable (\textbf{Fixed}) and stochastic (\textbf{Uniform}) settings. Conversely, the mechanisms in FFM and SHM, while effective for static patterns, are not sufficiently flexible to adapt to the variability of the uniform mode. This suggests their rewriting capability is effective but limited to predictable scenarios. Also, SHM results regarding the number of timesteps required to complete episodes may be indicative of brute force, although stable successful execution of $n-1$ corridors in some tasks may refute this.

\begin{figure}
    \centering
    \includegraphics[width=1.0\linewidth]{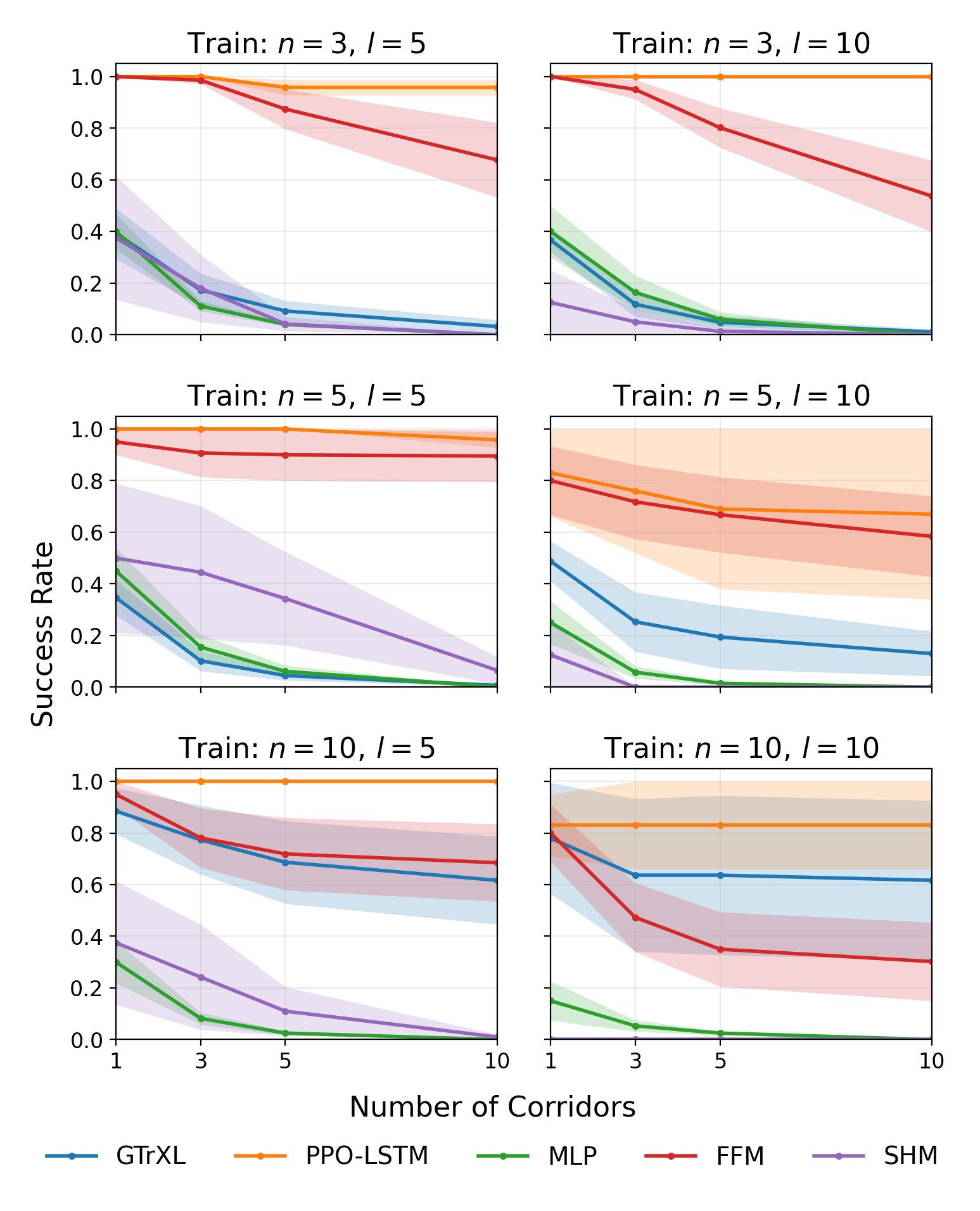}
    \vspace{-25pt}    
    \caption{Baseline comparison on \texttt{Endless T-Maze} under interpolation and extrapolation conditions, that is, where corridor lengths and fixed sampling are the same during training and validation, and in each validation task the parameter of the number of corridors varies in accordance with the experiment protocol. The result for Success Rate is mean$\pm$sem.}
    \Description{A line chart comparing baseline success rates on Endless T-Maze across interpolation and extrapolation settings, varying the number of corridors while keeping training and validation regimes consistent.}
    \label{fig:n-variations}
\end{figure}

\paragraph{\textbf{RQ3: What memory mechanisms can demonstrate generalization in memory rewriting?}}
When considering these research question, we consider the terms ``interpolation'' and ``extrapolation'' in context of \texttt{Endless T-maze} results. Interpolation means that the agent was able to cope with the validation task where the number of corridors or their length was less than the training configuration, while Extrapolation means the opposite result -- the number of corridors or their length in the validation task was greater.

The results in the Figure~\ref{fig:n-variations} show that PPO-LSTM demonstrated interpolation for all lengths in non-trivial training configurations with uniform mode, achieving complete success regardless of the number of corridors trained and validated. In addition, PPO-LSTM was able to demonstrate incomplete but successful extrapolation to a corridor length of 10, again regardless of the number of corridors with a non-trivial uniform 5 training task.

FFM, SHM, and GTrXL performed successfully in non-trivial fixed training configurations, the models show generalization to other numbers of corridors. The FFM agent demonstrated both extrapolation and interpolation across all such configurations, while SHM and GTrXL demonstrated only interpolation, and the transformer architecture performs better.

It is worth considering how the attention mechanism performed under uniform conditions using the example of the GTrXL baseline results. Agent managed to interpolate one uniform training task to fixed configuration task by the number of corridors. In particular, the training configuration $n = 3, \ l = 5, \ \text{regime} = \text{Uniform}$ showed a slight extrapolation to the task with 5 corridors with success $0.38\pm0.21$. 

The generalization results allow us to establish a clear hierarchy of flexibility among the memory mechanisms.

At the top is PPO-LSTM, which learns an abstract and adaptable strategy, enabling it to generalize successfully even in uniform environments. Below it lies FFM, which demonstrates a more limited and conditional form of flexibility. Its ability to both interpolate and extrapolate, but exclusively within the predictable confines of the Fixed mode, suggests it can generalize a learned pattern, but cannot adapt the pattern itself. At the bottom of this hierarchy is SHM, which proves to be the most rigid. Its ability is restricted to interpolation only, indicating that its learned strategy is tightly coupled with the specific parameters of the training environment and cannot be scaled even in predictable settings.

This distinct ranking of adaptability provides the piece of evidence, suggesting that the key to success lies in a fundamental property of the memory mechanism that dictates how it acquires and applies knowledge in new situations.

\paragraph{\textbf{RQ4: Which memory mechanisms are adaptive to efficiently ranking of rewritten information?}}
After the first sub-episode, a new random target is chosen in \texttt{Medium} and \texttt{Extreme} modes.  
To successfully complete the second and subsequent sub-episodes, the agent must remember and record all teleportations that have occurred, and in \texttt{Extreme} mode, it must also make logical conclusions in order to correctly recognize which colors correspond to which coordinates. In other words, we also test the ability not only to adaptively overwrite and forget old information, but also dynamically update the storage, reevaluating the relevance of all stored data in order to be able to focus on the current task.

Under these conditions, all baselines showed zero success Table~\ref{color-cubes-results}, which may indicate that their memory mechanisms are not sensitive to tasks where simply forgetting information does not help.

\begin{figure}
    \centering
    \includegraphics[width=1.0\linewidth]{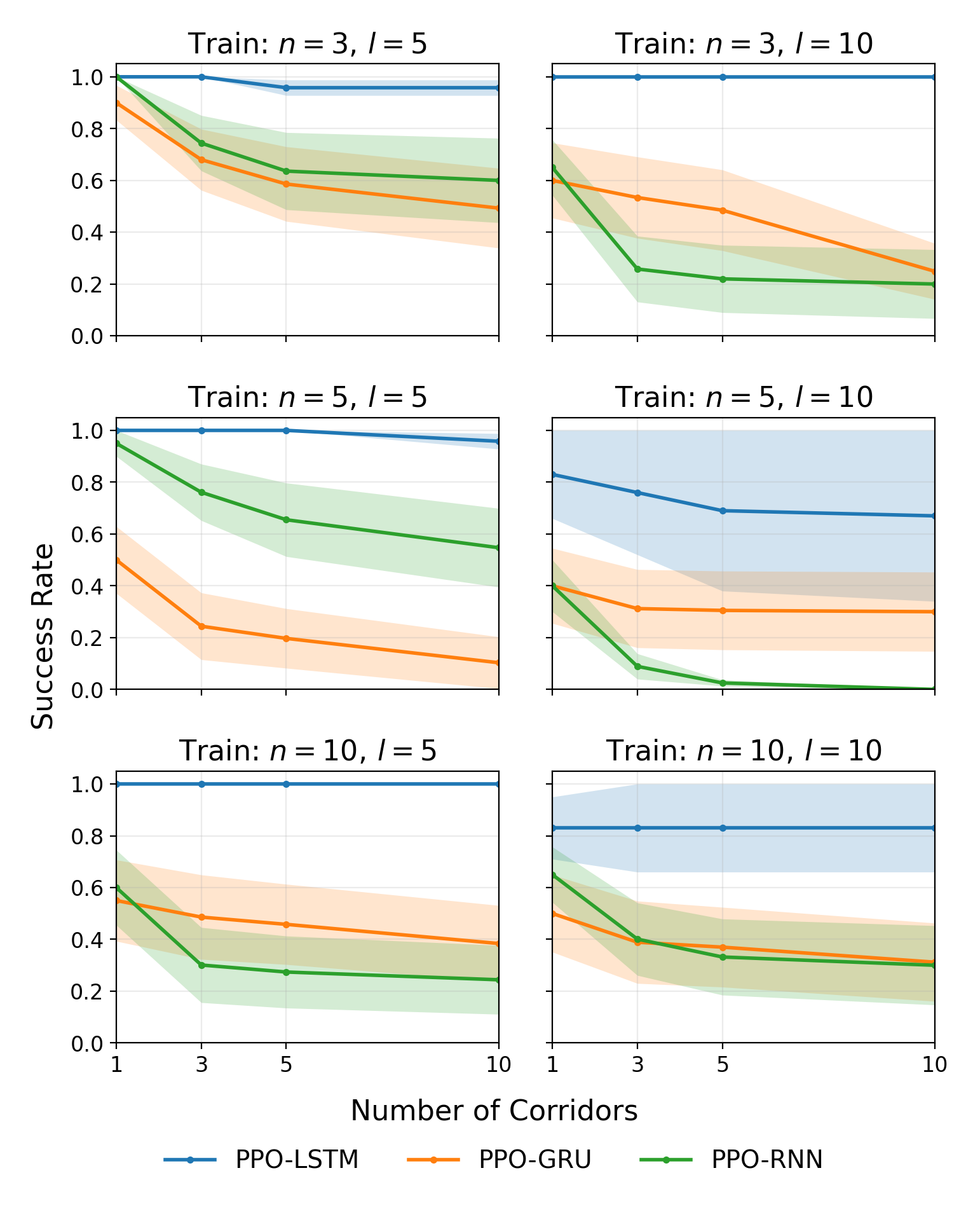}
    \vspace{-25pt}    
    \caption{Representatives of recurrent models comparison on \texttt{Endless T-Maze} under interpolation and extrapolation conditions, that is, where corridor lengths and fixed sampling are the same during training and validation, and in each validation task the parameter of the number of corridors varies in accordance with the experiment protocol. The result for Success Rate is mean$\pm$sem.}
    \Description{A plot comparing recurrent-model variants on Endless T-Maze across interpolation and extrapolation settings, reporting mean success rate with standard error.}
    \label{fig:ablation}
\end{figure}

\section{Ablation study}
Since the PPO-LSTM model outperformed other baselines under considered scenarios, a hypothesis was put forward that its effectiveness may be due to the specifics of its internal architecture, namely the presence of gates that separate and control different aspects of calculations.

To test this hypothesis, a series of comparative experiments was conducted, including modifications of PPO with alternative recurrent architectures -- PPO-RNN and PPO-GRU. The PPO-RNN model, based on a classic recurrent layer without gating mechanisms, will help in analyzing the general role of gates in memory rewriting tasks. Also PPO-GRU is an intermediate option in which the architecture is simplified compared to LSTM: the cell and hidden states are combined, and the number of gates is reduced from three (forget, input, output) to two (reset and update). This comparison allows us to evaluate not only the usefulness of gating as such, but also the impact of specific memory and information flow control mechanisms on overall performance.

For performance comparison, we used scenarios from experiments in the Endless T-Maze environment in fixed mode, with corridor lengths varying between 5 and 10, and the number of corridors varying between 3, 5, and 10. These scenarios were chosen because the extrapolation and interpolation conditions in them was useful in a comparative analysis of the success of the previous baselines taken.

The results on Figure~\ref{fig:ablation} show that RNN succeeds only in a limited subset of interpolation and extrapolation settings precisely those where SHM and FFM (also without gates) previously achieved partial success. At the same time, PPO-GRU demonstrated the ability to interpolate and extrapolate not only when changing the number of corridors in scenarios where SHM and FFM showed results, but also demonstrated a limited ability to generalize when changing the corridor length. However, PPO-GRU was inferior to PPO-LSTM in terms of generalization success over lengths.

A comparison of PPO-RNN as a representative of recurrent models without gating with PPO-GRU and PPO-LSTM, which showed significantly higher results, indicates that gating mechanisms have a key influence on the success of rewriting tasks. The higher performance of PPO-GRU over SHM, FFM, and PPO-RNN in its ability to generalize the solution to length variation further confirms the usefulness of gates in such tasks. At the same time, the advantage of PPO-LSTM over PPO-GRU in length generalization scenarios demonstrates the importance of an adaptive forgetting mechanism.

\begin{table}[t]
\centering
\small
\caption{The results in the all \texttt{Color-Cubes}. Medium and Extreme mode core parameters are: cubes = 3, sub-episodes = 3, grid size = 5, teleport chance = 0.3.}
\begin{adjustbox}{width=1\columnwidth}
\begin{tabular}{lccccc}
\toprule
 & \textbf{PPO-LSTM} 
 & \textbf{GTrXL} 
 & \textbf{SHM} 
 & \textbf{FFM} 
 & \textbf{MLP} \\
\midrule
 Trivial & \colresult{0.52}{0.10}{2}  & \colresult{1.00}{0.00}{1}  & \colresult{1.00}{0.00}{1} & \colresult{1.00}{0.00}{1} & 0.00$\pm$ 0.00\\
 Medium & 0.00$\pm$0.00  & 0.00$\pm$0.00  & 0.00$\pm$0.00 & 0.00 $\pm$0.00 & 0.00$\pm$ 0.00\\
 Extreme& 0.00$\pm$0.00  & 0.00$\pm$0.00  & 0.00$\pm$0.00 & 0.00 $\pm$0.00 & 0.00$\pm$ 0.00\\
\midrule
\end{tabular}
\end{adjustbox}
\label{color-cubes-results}
\end{table}

\section{Discussion}
The results from the preceding research questions establish a consistent performance hierarchy among the tested architectures. Across tasks involving information retention (RQ1), rewriting (RQ2), and generalization (RQ3), the observed ranking was: PPO-LSTM, FFM, SHM, GTrXL, and MLP.

This ranking correlates with the architectural approach each model takes to managing outdated information -- specifically, with its mechanism for forgetting. An analysis of the architectures reveals a trend:
\begin{itemize}
    \item \textbf{PPO-LSTM}, the top-performing model, utilizes a dedicated forget gate with learnable parameters. This allows for an adaptive, context-dependent approach to information retention and removal.
    \item \textbf{FFM} architecture implements a more defined, rule-based forgetting process where older information is systematically replaced. This approach was effective in predictable fixed environments but was associated with a significant performance degradation in uniform settings.
    \item \textbf{GTrXL} as mentioned earlier, the model is vulnerable to sparse reward environments, but at the same time it was able to show generalization in some scenarios. Nevertheless, the stabilization of architecture learning and information writing to the cache occurs with the help of gating, which once again emphasizes their necessity.
    \item \textbf{SHM} has a similar concept of matrix memory structuring to FFM, but lacks an explicit forgetting mechanism. SHM has a dynamic update of information significance. This design correlated with lower performance compared to FFM, particularly in generalization tasks.
\end{itemize}

Based on this observed correlation, we hypothesize that the nature and adaptability of the forgetting mechanism may be key factors determining an agent's effectiveness in performing tasks that require memory rewriting in dynamic environments.

A comparison in an ablation study with PPO-RNN confirmed that PPO-LSTM gates contributed to its success, while a comparison of its gates with PPO-GRU gates highlighted the need for an adaptive forgetting mechanism.

However, PPO-LSTM performed worse than other memory-based baselines in Trivial Color-Cubes. Since this case study does not involve rewriting, it cannot be said that SHM, FFM, and GTrXL are superior, but PPO-LSTM may not be robust to grid environments.

\section{Limitations \& Future Work}
One of the key limitations of this study is the small number of scientific papers devoted directly to the mechanisms of memory rewriting in RL. This complicates the comparison of the results obtained. In addition, a broader range of benchmarks and a larger number of diverse agents are needed to more accurately assess and isolate individual memory abilities and mechanisms. Finally, based on the analysis of the results obtained, a promising direction for further work is the development of a proprietary mechanism that specifically addresses the task of memory rewriting.

\section{Conclusion}
Our study reveals that memory retention alone is insufficient for solving RL tasks that require continual adaptation under partial observability.  
Through the proposed \texttt{Endless T-Maze} and \texttt{Color -Cubes} benchmarks, we isolate and evaluate the ability of RL agents to rewrite memory -- to selectively discard obsolete information and integrate new evidence as environments evolve.  
Across recurrent, transformer-based, and structured-memory architectures, we observe a consistent hierarchy of adaptive competence: agents with explicit, learnable forgetting mechanisms (such as LSTM-based policies) exhibit strong performance and generalization, whereas attention- and cache-based models struggle when rewriting is essential rather than optional.
These findings suggest that the core challenge in memory-intensive decision-making is not the preservation of information but its controlled transformation over time. 
Future memory systems should therefore move beyond static representations of history toward architectures that model memory as an active, continuously updated belief state -- capable of forgetting, rewriting, and reallocating representational capacity as conditions change.  
We view this as a step toward closing the gap between artificial and biological memory, where adaptation and forgetting are not failure modes but fundamental features of intelligent behavior.

\section*{Acknowledgments}
Some figures in this paper use Chrome Dinosaur–style sprites obtained from Wikimedia Commons, which are licensed under their respective Creative Commons licenses (e.g., CC BY~4.0 / CC BY-SA~4.0). We gratefully acknowledge the creators and maintainers of these assets.



\bibliographystyle{ACM-Reference-Format} 
\bibliography{sample}


\appendix
\section{Detailed Environments Descriptions}
\label{app:sec:envs}
\subsection{Detailed Description: Endless T-Maze.}
\label{app:t-maze}
The \texttt{Endless T-Maze} environment extends the classical T-Maze~\citep{ni2023transformers} into a continual, multi-stage task that explicitly tests an agent's ability to rewrite memory as new cues arrive.  
The environment consists of a sequential chain of $N$ T-shaped corridors, each ending in a left–right junction.  
At the beginning of every corridor, the agent receives a binary cue represented as a two-dimensional vector -- $(1, 0)$ for ``turn left'' or $(0, 1)$ for ``turn right''.  
This cue is displayed only at the corridor's start and becomes invalid once the next corridor begins, forcing the agent to overwrite previously stored information rather than accumulate it.  

During traversal, the agent moves forward along the corridor, observing its normalized position (a scalar between $0$ and $1$) and the cue vector, which becomes $(0, 0)$ after the first step.  
The observation space therefore consists of three values: $(\text{cue}_1, \text{cue}_2, \text{position})$.  
The discrete action space includes \texttt{MOVE\_\allowbreak FORWARD}, \texttt{TURN\_\allowbreak LEFT}, and \texttt{TURN\_\allowbreak RIGHT}.  
A correct turn at a junction yields a reward of $+1$, an incorrect turn results in a penalty of $-1$, and each idle or forward step incurs a small penalty of $-0.01$ to encourage efficient movement.

Two corridor-length regimes are supported:
\begin{itemize}
    \item \textbf{Fixed:} all corridors have the same predefined length $l_{\max}$.
    \item \textbf{Uniform:} corridor lengths are sampled from a uniform distribution $l \sim \mathcal{U}[1, l_{\max}]$, introducing variability in memory duration between cue presentation and decision.
\end{itemize}
Although the environment is conceptually infinite, in practice episodes are truncated after a fixed number of corridors or a global step limit to maintain stable training and evaluation.  
This setup directly measures an agent's capacity to discard obsolete cues and replace them with new, task-relevant information in a continually evolving sequence of decisions.

\subsection{Detailed Description: Color-Cubes.}
\label{app:cubes}
The environment consists of a two-dimensional grid of size $G{\times}G$ with $N$ cubes placed at random, unique positions.  
Each cube is assigned a unique color $c \in \{0, 1, \dots, N{-}1\}$, and the agent's goal is to sequentially locate and interact with cubes of a specified target color.

An episode is divided into $K$ sub-episodes.  
At the start of each sub-episode, the agent receives the target color $c_{\text{target}}$.  
The agent must navigate to the corresponding cube and execute the \texttt{INTERACT} action.  
A successful interaction concludes the sub-episode, grants a reward of $+1.0$, teleports the collected cube to a new random unoccupied cell, and assigns a new target color from the remaining set of cubes.  
The agent also receives a new \texttt{full\_\allowbreak state\_\allowbreak update}.  
At every timestep where no successful interaction occurs, a random non-target cube teleports with probability $p_{\text{teleport}}$, triggering another \texttt{full\_\allowbreak state\_\allowbreak update}.  
If neither event happens, the update is withheld, forcing the agent to act using only its internal memory.  
Because the world state may change without notice, outdated memory can lead the agent to an empty cell where a cube previously stood.

\newpage
\textbf{Difficulty variants.}  
\begin{itemize}
    \item \texttt{\textbf{Trivial:}} One cube and one target ($N{=}K{=}1$). The agent only needs to remember and interact with a single target cube.
    \item \texttt{\textbf{Medium:}} The standard setup with multiple cubes and complete state updates (positions and colors).
    \item \texttt{\textbf{Extreme:}} Teleportation updates omit color identifiers. The agent must recall the initial color–position mapping, identify which cube moved, and update its internal representation accordingly.
\end{itemize}

An episode terminates after $K$ successful interactions or upon reaching the time limit.  
This setup explicitly tests whether agents can maintain, detect, and rewrite memory representations when the environment changes unpredictably.

\section{Training Configuration}
\label{app:training}

\begin{table}[htbp]
\centering
\caption{Hyperparameters for SHM.}
\label{tab:hyperparams_shm}
\begin{tabular}{lc}
\toprule
\textbf{Hyperparameter} & \textbf{Value} \\
\midrule
Hidden size $(h)$ & 512 \\
Memory size $(m)$ & 128 \\
Post-processing size & 1024 \\
Learning rate & $5 \times 10^{-5}$ \\
Discount $(\gamma)$ & 0.99 \\
GAE lambda $(\lambda)$ & 1.0 \\
Gradient clipping & 0.5 \\
Entropy coefficient & 0.001 \\
Value loss coefficient & 0.5 \\
BPTT length & 1024 \\
Train batch size & 65536 \\
Minibatch size & 8192 \\
PPO epochs & 6 \\
Number of workers & 12 \\
Total timesteps & 2M \\
Framework & Ray RLlib \\
\bottomrule
\end{tabular}
\end{table}

\begin{table}[htbp]
\centering
\caption{Hyperparameters for FFM.}
\label{tab:hyperparams_ffm}
\begin{tabular}{lc}
\toprule
\textbf{Hyperparameter} & \textbf{Value} \\
\midrule
Hidden size & 128 \\
Memory size $(m)$ & 128 \\
Post-processing size & 1024 \\
Learning rate & $5 \times 10^{-5}$ \\
Discount $(\gamma)$ & 0.99 \\
GAE lambda $(\lambda)$ & 1.0 \\
Gradient clipping & 0.5 \\
Entropy coefficient & 0.001 \\
Value loss coefficient & 0.5 \\
BPTT length & 1024 \\
Train batch size & 65536 \\
Minibatch size & 8192 \\
PPO epochs & 6 \\
Number of workers & 8 \\
Total timesteps & 2M \\
Framework & Ray RLlib \\
\bottomrule
\end{tabular}
\end{table}

\begin{table}[htbp]
\centering
\caption{Hyperparameters for MLP.}
\label{tab:hyperparams_mlp}
\begin{tabular}{lc}
\toprule
\textbf{Hyperparameter} & \textbf{Value} \\
\midrule
Hidden size & 256 \\
Memory & None \\
Learning rate & $5 \times 10^{-5}$ \\
Discount $(\gamma)$ & 0.99 \\
GAE lambda $(\lambda)$ & 1.0 \\
Gradient clipping & 0.5 \\
Entropy coefficient & 0.001 \\
Value loss coefficient & 0.5 \\
BPTT length & 1024 \\
Train batch size & 65536 \\
Minibatch size & 8192 \\
PPO epochs & 6 \\
Number of workers & 2 \\
Total timesteps & 2M \\
Framework & Ray RLlib \\
\bottomrule
\end{tabular}
\end{table}

\begin{table}[htbp]
\centering
\caption{Hyperparameters for GTrXL.}
\label{tab:hyperparams_gtrxl}
\begin{tabular}{lc}
\toprule
\textbf{Hyperparameter} & \textbf{Value} \\
\midrule
Hidden size & 512 \\
Memory length & 100 \\
Transformer blocks & 4 \\
Attention heads & 8 \\
Embedding dimension & 512 \\
Positional encoding & Relative \\
GTrXL gating & Enabled \\
Learning rate (initial) & $2.75 \times 10^{-4}$ \\
Learning rate (final) & $1 \times 10^{-5}$ \\
Discount $(\gamma)$ & 0.995 \\
GAE lambda $(\lambda)$ & 0.95 \\
Gradient clipping & 0.25 \\
Entropy coeff (initial) & 0.001 \\
Entropy coeff (final) & $1 \times 10^{-6}$ \\
Value loss coefficient & 0.5 \\
PPO epochs & 3 \\
Number of workers & 32 \\
Steps per worker & 512 \\
Minibatches per epoch & 8 \\
Total timesteps & 2M \\
Framework & Custom PPO \\
\bottomrule
\end{tabular}
\end{table}

\begin{table}[htbp]
\centering
\caption{Hyperparameters for PPO-LSTM.}
\label{tab:hyperparams_lstm}
\begin{tabular}{lc}
\toprule
\textbf{Hyperparameter} & \textbf{Value} \\
\midrule
LSTM hidden size & 128 \\
Actor network & [128, 128] \\
Critic network & [128, 128] \\
Learning rate (initial) & $3 \times 10^{-4}$ \\
Learning rate (final) & $1 \times 10^{-5}$ \\
LR schedule & Linear \\
Discount $(\gamma)$ & 0.99 \\
GAE lambda $(\lambda)$ & 0.98 \\
Gradient clipping & 0.5 \\
Entropy coefficient & 0.1 \\
Value loss coefficient & 0.5 \\
Sequence length & 128 \\
Minibatch size & 512 \\
PPO epochs & 10 \\
Number of environments & 16 \\
Steps per environment & 128 \\
Total timesteps & 2M \\
Framework & Stable-Baselines3 \\
\bottomrule
\end{tabular}
\end{table}

\clearpage
\end{document}